\documentclass{article}

\usepackage{PRIMEarxiv}

\usepackage[utf8]{inputenc} 
\usepackage[T1]{fontenc}    
\usepackage{amsmath}        
\usepackage{amssymb}        
\usepackage{hyperref}       
\usepackage{url}            
\usepackage{booktabs}       
\usepackage{multirow}       
\usepackage{nicefrac}       
\usepackage{microtype}      
\usepackage{fancyhdr}       
\usepackage{graphicx}       
\usepackage{subcaption}     
\graphicspath{{figures/}{./}}

\newcommand{\Pcog}{P_{\mathrm{cognition}}}
\newcommand{\Pcode}{P_{\mathrm{coding}}}
\newcommand{\Pexp}{P_{\mathrm{expert}}}
\newcommand{\Pagent}{P_{\mathrm{agent}}}
\newcommand{\Pcomp}{P_{\mathrm{compiler}}}
\newcommand{\fcog}{f_{\mathrm{cognition}}}
\newcommand{\fcode}{f_{\mathrm{coding}}}
\newcommand{\Hhat}{\hat{H}}

\title{Cognitive-YOLO: LLM-Driven Architecture Synthesis from First Principles of Data for Object Detection}

\author{
  Jiahao Zhao \\
  School of Computer Science and Technology \\
  Xi'an University of Posts and Telecommunications \\
  \texttt{zjh@stu.xupt.edu.cn} \\
}

\begin{document}
\maketitle

\begin{abstract}
General-purpose object detection algorithms face a domain-shift bottleneck when deployed in vertical scenarios, and manually customizing a network architecture relies heavily on expert experience with high trial-and-error cost. To address these problems, we present \textbf{Cognitive-YOLO}, an adaptive object-detection model synthesis system driven by a large language model (LLM) collaborating with an autonomous agent. Cognitive-YOLO builds a fully automated \emph{Analyze--Synthesize--Compile} pipeline. First, a data-profiling module quantitatively extracts dataset features, guiding a ReAct-paradigm agent to autonomously retrieve matching network modules from a curated component library. Second, the LLM performs logical reasoning and topological assembly conditioned on the data features, producing a specification-compliant neural architecture description. Finally, a backend hybrid compiler dynamically instantiates the model, closing the loop with end-to-end training and evaluation scheduling. Cross-validation on datasets including rail-surface defects and rice diseases shows that the lightweight models synthesized by Cognitive-YOLO markedly compress the parameter count (down to 1.9M) while matching or surpassing mainstream baselines such as YOLOv11n in mAP@0.5:0.95, effectively balancing model compactness and feature-representation capability.
\end{abstract}


\section{Introduction}
In recent years, general-purpose object detection algorithms represented by the YOLO family~\cite{Redmon_Divvala_Girshick_Farhadi_2015} have achieved breakthrough progress in computer vision~\cite{vijayakumar2024yolo}. However, as deep learning accelerates its deployment into real physical-world vertical scenarios---such as tiny-object detection from UAV viewpoints and defect recognition in complex industrial environments---general pre-trained models often encounter a fatal bottleneck: \emph{domain shift}. Because the data characteristics of standard benchmarks differ enormously from those of actual deployment environments, general models frequently suffer from significantly degraded generalization and severely insufficient domain adaptability in practice~\cite{fan2023towards}.

Moreover, the data distributions of real industrial and specific vertical scenarios are usually strongly long-tailed~\cite{zhang2023reconciling}, which means that edge samples containing rare poses, extreme occlusion, or special illumination occupy a very large proportion. Every concrete scenario has its own unique physical distribution law. This determines that the key to solving the pain points of complex-scene detection is no longer to blindly pursue a single theoretically universal model, but to tailor, for each scenario-specific dataset, a network architecture that best fits its physical prior characteristics. This paradigm evolution from \emph{general-model adaptation} toward \emph{data-driven customization} carries important research value and engineering significance.

This paper aims to design and implement an adaptive object-detection system based on a large language model (LLM). The system, which we call \textbf{Cognitive-YOLO}, provides flexible visual-perception support for complex and diverse vertical applications (e.g., industrial defect inspection and security monitoring). Cognitive-YOLO first generates a structured profile by extracting the quantitative features of a dataset, and then leverages the collaboration of an LLM and an agent to retrieve and dynamically assemble the network architecture that best matches the current data characteristics. By deeply analyzing the data-distribution features, the system automatically generates a lightweight detection model suitable for a specific scenario, providing an efficient architecture-customization scheme for domain-specific detection tasks and effectively simplifying the model-design process that traditionally depends heavily on manual trial and error.

The contributions of this paper are as follows:
\begin{enumerate}
    \item We propose Cognitive-YOLO, a novel paradigm for object-detection architecture synthesis that reasons from the first principles of the dataset, and we formalize the cognitive process that maps data features to an optimal architecture.
    \item We design and implement a flexible, decoupled \emph{Analyze--Synthesize--Compile} framework in which a ReAct-paradigm agent externalizes the latent expert design intuition into an observable, structured retrieval trajectory.
    \item We validate the effectiveness of Cognitive-YOLO on four cross-domain datasets. Experimental results show that Cognitive-YOLO generates lightweight models whose comprehensive performance matches or surpasses mainstream baselines such as YOLOv11n while significantly compressing the parameter count.
\end{enumerate}

\section{Related Work}
\subsection{Object Detection}
Single-stage object detection algorithms represented by the YOLO family have been widely applied in industrial and real physical scenarios in recent years. After many iterations, modern YOLO networks have precipitated an extremely clear and rigorous topological paradigm whose core architecture is strictly decoupled into three standard components: a Backbone, a Neck, and a detection Head. The Backbone deeply integrates cross-stage partial networks with attention mechanisms, efficiently extracting multi-level features from shallow textures to deep semantics while substantially reducing floating-point operations. The Neck builds a bidirectional information-flow mechanism relying on spatial pyramid pooling and path-aggregation networks, realizing the cross-scale deep integration of shallow localization information and deep semantic information. The detection Head has evolved toward an anchor-free and decoupled-head mechanism, which not only removes the tedious prior-anchor clustering process but also effectively resolves the underlying conflict between the classification and regression tasks in terms of spatial feature alignment.

\subsection{LLM Reasoning and Agent Technology}
Large language models, based on the Transformer architecture and the self-attention mechanism~\cite{vaswani2017attention}, internalize massive general world knowledge and algorithmic topological norms through autoregressive causal language modeling on large-scale unlabeled corpora. To break through the probabilistic-generation limitations of traditional models on complex mathematical calculus and multi-step logical reasoning, the research community introduced Chain-of-Thought (CoT)~\cite{Wei_Wang_Schuurmans_Bosma_Ichter_Xia_Chi_Le_Zhou_2022}, enabling models to generate explicit intermediate reasoning steps before producing the final answer, thereby decomposing complex problems into tractable sub-modules. In particular, a new generation of reasoning models represented by DeepSeek-R1~\cite{DeepSeek-AI_Guo_Yang_Zhang_Song_Zhang_Xu_Zhu_Ma_Wang_et_al_2025} introduces large-scale test-time computation and, relying on reinforcement-learning algorithms such as Group Relative Policy Optimization~\cite{shao2024deepseekmath}, shifts the optimization objective from fitting human subjective preference toward pursuing objective logical correctness. This paradigm leap from intuitive content generation to systematic logical deduction provides a powerful underlying cognitive support for the automated design of deep-learning architectures.

In engineering practice, Agentic Retrieval-Augmented Generation (Agentic RAG)~\cite{singh2025agenticretrievalaugmentedgenerationsurvey} further breaks the closed-loop computation limitation of LLMs, endowing them with the ability to interact with external digital environments. By introducing multi-agent collaboration paradigms such as Reasoning-and-Acting (ReAct)~\cite{yao2023reactsynergizingreasoningacting}, a system can guide the model to generate internal reasoning paths before taking actions and to autonomously invoke external tools such as SQL database queries. This mechanism successfully externalizes the implicit algorithm-design path, which originally belongs to human experts, into observable and structured component-retrieval and self-correction trajectories.

\subsection{LLM-Assisted Model Design}
\textbf{LLM-assisted evolution and code optimization.} Yu and Zutty et al.~\cite{yu2025llmge} proposed the LLM-Guided Evolution (LLM-GE) framework. Unlike traditional NAS that uses fixed mutation operators, this work uses an LLM to directly modify the source code of a YOLO model and establishes a feedback loop through an Evolution-of-Thought (EoT) mechanism, allowing the model to learn from previous performance feedback and iteratively optimize the architecture. Experiments show that this method significantly improves detection accuracy on the KITTI dataset.

\textbf{LLM-driven dynamic search.} To address the low search efficiency of NAS, Kong et al.~\cite{kong2025phasenas} proposed the PhaseNAS framework. This work uses an LLM to structurally describe architecture templates and introduces a dynamic phase-adaptation strategy that switches between ``broad exploration'' and ``fine-grained optimization'' according to real-time scores. This method successfully reduced the search time of YOLOv8 variants by 86\%, demonstrating the great potential of LLMs in accelerating architecture search.

\textbf{Low-data adaptability of multimodal large models.} Beyond assisting architecture design, Elamon and Davoudi~\cite{elamon2025beyond} explored the direct application of multimodal large models (LMMs) under low-data regimes. The study found that, in few-shot scenarios, a fine-tuned multimodal LLM exhibits stronger generalization and contextual reasoning ability than traditional CNNs.

In summary, the field is in a critical period of transition from traditional NAS toward LLM-driven intelligent design. Although existing LLM-NAS works have made preliminary progress in code generation and search efficiency, most methods still treat the LLM as a ``mutator'' or a ``surrogate model'' within an optimization loop, lacking research that starts from data features (e.g., scale and density) and directly leverages the LLM's logical reasoning ability for end-to-end architecture synthesis. This work is based on precisely this background, aiming to explore a new data-driven LLM architecture-synthesis method.

\section{Method: Cognitive Modeling of Architecture Synthesis}
In real engineering scenarios, designing the optimal network architecture from a raw dataset is not a simple end-to-end mathematical formula, but undergoes a complex cognitive transformation process. We formalize this process as follows. First, given a task-specific dataset $D$, the system needs to quantify its physical features into an explicit meta-feature vector through a feature-extraction operator $\phi$, i.e.,
\begin{equation}
    M = \phi(D),
    \label{eq:meta}
\end{equation}
which contains the object-scale distribution, scene density, and so on. In the design process of a human expert, after observing the data features $M$, the expert does not immediately write code, but instead generates in their mind an \emph{expert-heuristic latent state}, denoted $H$. This latent state $H$ contains highly nonlinear design intuition and matching rules (for example: when the proportion of small objects exceeds 30\% and the compute budget is limited, a BiFPN module that preserves high-resolution features should be introduced in the Neck). Subsequently, the expert combines the known SOTA component library $K$ to instantiate the latent state $H$ into a concrete network architecture $A^{*}$. The traditional human design path can be expressed as
\begin{equation}
    H = \fcog(M),
    \label{eq:cog}
\end{equation}
\begin{equation}
    A^{*} = \fcode(H, K).
    \label{eq:code}
\end{equation}
Combining Eq.~\eqref{eq:cog} and Eq.~\eqref{eq:code} yields
\begin{equation}
    A^{*} = \fcode\!\big(\fcog(M),\, K\big).
    \label{eq:combine}
\end{equation}

The process by which a human expert designs the optimal architecture $A^{*}$ from the data features $M$ and the module library $K$ is actually a generative process containing the latent variable $H$, whose truly observed expert marginal distribution is expressed as an integral over all possible latent states:
\begin{equation}
    \Pexp(A^{*}\mid M, K) = \int \Pcode(A^{*}\mid H, K)\,\Pcog(H\mid M)\,\mathrm{d}H.
    \label{eq:marginal}
\end{equation}
In Eq.~\eqref{eq:marginal}, $\Pcog$ models the probability of the expert's cognitive mapping from data to design intuition, while $\Pcode$ models the instantiation probability from intuition to a concrete coded architecture.

The architecture generation of an LLM is a conditional-probability autoregressive process based on parameters $\theta$, denoted $P_{\theta}(A\mid M, K)$. During pre-training, the LLM absorbs massive amounts of open-source deep-learning code and top-tier academic papers. In a physical sense, these corpora are exactly the observation samples independently and identically distributed (i.i.d.) sampled by human experts from the true distribution $\Pexp$ in the past. According to maximum-likelihood estimation theory, the goal of LLM pre-training is to find the optimal parameters $\theta^{*}$ that minimize the Kullback--Leibler (KL) divergence between the true distribution and the model-predicted distribution:
\begin{equation}
    \theta^{*} = \arg\min_{\theta}\, D_{\mathrm{KL}}\!\big(\Pexp(A\mid M, K) \,\|\, P_{\theta}(A\mid M, K)\big).
    \label{eq:kl}
\end{equation}
Expanding the KL divergence using the logarithm property gives
\begin{equation}
    D_{\mathrm{KL}}\!\big(\Pexp \,\|\, P_{\theta}\big)
    = \mathbb{E}_{A\sim\Pexp}\!\big[\log \Pexp(A\mid M, K)\big]
    - \mathbb{E}_{A\sim\Pexp}\!\big[\log P_{\theta}(A\mid M, K)\big].
    \label{eq:klexpand}
\end{equation}
The first term is the negative entropy of the true data distribution, which is independent of the model parameters $\theta$ and is therefore a constant. Hence, minimizing the KL divergence is strictly equivalent to maximizing the second term (i.e., minimizing the negative log-likelihood). As empirical-risk minimization converges, the KL divergence tends to a minimum and the model distribution infinitely approaches the true expert distribution:
\begin{equation}
    P_{\theta}(A\mid M, K) \approx \Pexp(A\mid M, K).
    \label{eq:converge}
\end{equation}
Substituting the expert latent-variable integral model of Eq.~\eqref{eq:marginal} into the above yields the core equivalent fitting equation of the system:
\begin{equation}
    P_{\theta}(A\mid M, K) \approx \int \Pcode(A\mid H, K)\,\Pcog(H\mid M)\,\mathrm{d}H.
    \label{eq:fitting}
\end{equation}

Although the above analysis theoretically proves that an LLM possesses the ability to generate the optimal network architecture, in practical engineering, directly relying on a single LLM's forward propagation for end-to-end generation faces fatal defects. The huge SOTA component library $K$ far exceeds the LLM's context-window limit, and the extremely complex integral operation is compressed into the LLM's black-box parameters, which easily produces \emph{hallucination}, causing the generated architecture to be physically broken and unusable in topology. The system should not let the large model blindly fit this unobservable integral equation, but should externalize the expert latent state $H$ into an observable, structured retrieval trajectory through a multi-agent collaboration mechanism.

\subsection{Externalizing Cognition via a ReAct Agent}
The system deploys a data-driven architect agent based on the ReAct paradigm. This agent does not directly generate code, but performs multi-round exploration in the database by executing a series of discrete tool calls (such as \texttt{sql\_query}). We define the reasoning trajectory formed by the agent over $T$ rounds of interaction, together with the finally selected module subset, as the \emph{explicit cognitive state} $\Hhat$:
\begin{equation}
    \Hhat = \{\tau_{1}, \tau_{2}, \dots, \tau_{T},\, K_{\mathrm{sub}}\},
    \label{eq:hhat}
\end{equation}
where $\tau_{t}$ denotes the reasoning trace and tool action at round $t$, and $K_{\mathrm{sub}}\subset K$ is the retrieved high-relevance module subset. This process actually transforms the originally black-box $\Pcog(H\mid M)$ into an observable sequential decision-making process:
\begin{equation}
    \Pagent(\Hhat\mid M, K) = \prod_{t=1}^{T} \pi_{\theta}\!\big(a_{t}\mid s_{t}, M\big),
    \label{eq:agent}
\end{equation}
where $a_{t}$ is the tool action taken at step $t$ and $s_{t}$ is the interaction state. This design bypasses the long-context limitation and forces the model to perform rigorous logical reasoning grounded in the dataset, effectively approximating the expert's early-stage investigation process.

After the explicit cognitive state $\Hhat$ is established, the originally large and vague knowledge base $K$ is reduced to a high-information-density subset $K_{\mathrm{sub}}$ together with its reasoning logic. At this point, the backend reasoning LLM takes over the task and only needs to focus on topological assembly and parameter alignment, instantiating $\Hhat$ into a neural architecture description:
\begin{equation}
    \Pcomp(A\mid M, \Hhat) \approx \Pcode(A\mid H, K).
    \label{eq:compile}
\end{equation}
In other words, the compiler's instantiation conditioned on the explicit state $\Hhat$ faithfully approximates the expert's coding process from the latent state $H$, while remaining fully observable and verifiable.

\section{System Implementation}
\begin{figure}[htbp]
   \centering
   \includegraphics[width=0.98\textwidth]{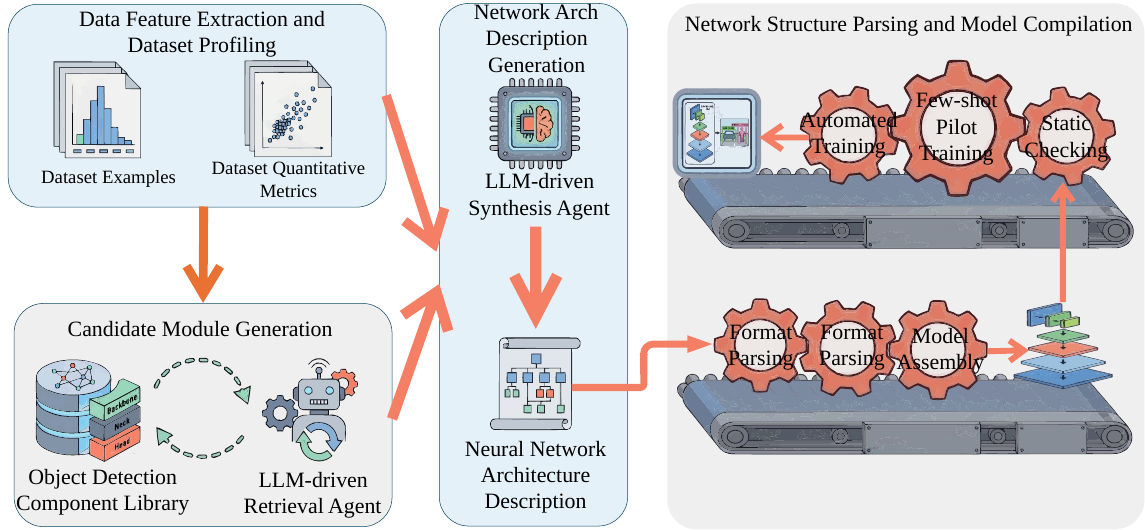}
   \caption{The \emph{Analyze--Synthesize--Compile} pipeline of Cognitive-YOLO. A data-profiling module first extracts quantitative meta-features from the dataset; a ReAct-paradigm agent then retrieves candidate modules from the SOTA component library; the reasoning LLM assembles them into a neural architecture description; and a backend compiler instantiates and validates the trainable model.}
   \label{fig:workflow}
\end{figure}

\subsection{Data Feature Extraction and Profile Construction}
The system must possess the ability to autonomously perceive the physical characteristics of the target dataset. To quantify the inherent distribution law of the dataset, the analysis stage first comprehensively traverses the user-uploaded raw images and annotation files through automated scripts. The specific extracted feature dimensions, their computation methods, and their meanings in guiding model design are shown in Table~\ref{tab:features}. These indicators transform a complex visual scene into structured data that is easy for the LLM to parse, providing a factual basis for the subsequent network-architecture reasoning.

\begin{table}[htbp]
\centering
\caption{Features generated by the data feature-extraction and profiling module.}
\label{tab:features}
\small
\begin{tabular}{@{}l l p{4.6cm}@{}}
\toprule
\textbf{Feature} & \textbf{Computation} & \textbf{Meaning for design} \\
\midrule
\texttt{num\_images} & image-file count & dataset scale; training iterations \\
\texttt{num\_classes} & number of class configs & task complexity; output-layer width \\
\texttt{classes} & list of class names & specific detection target types \\
\texttt{num\_objects} & total annotation boxes & data richness \\
\texttt{avg\_objects\_per\_image} & total objects / images & scene crowding; occlusion strategy \\
\texttt{class\_distribution} & per-class frequency & class balance; balancing strategy \\
\texttt{object\_size\_distribution} & small/medium/large area ratio & object-scale property; head design \\
\texttt{image\_size} & mean image width/height & input-resolution setting \\
\bottomrule
\end{tabular}
\end{table}

\subsection{Agent-Based Component Library and Intelligent Retrieval}
This stage is implemented by an agent with tool-calling capability. The agent first receives the analysis report output by the data-profiling module, which contains key feature indicators such as the number of classes, the object-scale distribution, and the scene crowding degree, and randomly selects some images from the dataset as samples. The agent then accesses the YOLO module database through SQL query statements. This database stores more than 300 curated module records, covering four major categories: Backbone feature-extraction networks, Neck feature-fusion networks, Head detection networks, and attention-mechanism modules.

In the module-screening process, the agent performs targeted matching according to the features in the data profile and the dataset samples. For example, when the proportion of small objects in the dataset exceeds a certain threshold, the model prioritizes modules with multi-scale feature-fusion capability; when class imbalance is obvious, it tends to select detection-head modules with class-balancing strategies. At the same time, the agent also reads the YAML configuration files corresponding to the candidate modules, analyzes their design ideas and implementation details, and finally screens out no fewer than 10 modules as the input for the next stage.

\subsection{Neural Architecture Description Generation}
This stage is undertaken by a large model with strong reasoning ability. The model first receives the candidate-module list from the previous stage, which contains key information such as each module's functional characteristics, applicable position, and design ideas. At the same time, the reasoning model reads the standard YOLO11 architecture template as a reference baseline, which defines the standard structure of the Backbone, Neck, and Head, as well as the connection rules between layers.

In the process of generating the neural architecture description, the reasoning model needs to comprehensively consider multiple factors. The first is the functional positioning of modules: the Backbone is usually responsible for high-level feature extraction and requires modules with strong feature-extraction capability; the Neck is responsible for multi-scale feature fusion and requires modules that can effectively integrate features of different levels; the Head is responsible for the final detection output and requires detection-head modules suitable for the detection task. The second is the matching of the number of channels between layers: the input and output channels of each layer must satisfy mathematical compatibility constraints, otherwise the framework will fail when loading the model. In addition, for some modules with special parameter requirements---for example, some modules only need to specify the output channels rather than the complete channel parameters---the reasoning model must strictly follow these specifications to avoid conflicts when the framework automatically injects parameters.

\subsection{Architecture Parsing and Model Compilation}
In the model-construction process, the configuration file first needs to be read and parsed, converting its layer definitions, connection relationships, and related parameters into structured information that the program can recognize. The system then looks up the corresponding implementation class in the registry according to the module name and completes the instantiation of each layer module in combination with the parameters. In this process, the program not only completes the layer-by-layer parsing of the network, but also further handles the input-output relationships between layers, the channel-number changes, and the encapsulation of the task type, thereby converting the static structural description into a complete trainable model. We call this process \emph{model compilation}, meaning that the network structure described in the configuration file is converted into a model object that can be directly used for training and inference.

After compilation is completed, the system performs a forward-propagation test on the generated model to verify whether the network structure can run normally and to check whether the dimension of the output result meets expectations. This step can discover structural-configuration problems in time before training begins, improving the reliability of the model-construction process. In this way, the definition, parsing, and compilation of the model structure are unified into the same process, making the overall implementation clearer and more suitable for subsequent experiments and model adjustment.

\section{Evaluation Metrics}
In object detection, an image may contain multiple objects, and the model outputs multiple bounding boxes. To classify the predicted boxes into TP, FP, or FN, a spatial-dimension evaluation metric is introduced: Intersection over Union (IoU). IoU is a geometric metric that quantifies the spatial overlap between the predicted bounding box $B_{p}$ generated by the model and the ground-truth bounding box $B_{gt}$, mathematically defined as the ratio of their intersection area to their union area:
\begin{equation}
    \mathrm{IoU} = \frac{\mathrm{Area}(B_{p}\cap B_{gt})}{\mathrm{Area}(B_{p}\cup B_{gt})}.
    \label{eq:iou}
\end{equation}

The evaluation metrics of object detection mainly include Precision, Recall, Accuracy, and F1-score. Most of these metrics are built on the confusion matrix (CM), which compares the classifier's predictions with the actual categories, as shown in Table~\ref{tab:cm}. The matrix is divided into true positives (TP), false negatives (FN), false positives (FP), and true negatives (TN).

\begin{table}[htbp]
\centering
\caption{Confusion matrix.}
\label{tab:cm}
\begin{tabular}{cc|cc}
\toprule
\multicolumn{2}{c|}{} & \multicolumn{2}{c}{\textbf{Predicted}} \\
\multicolumn{2}{c|}{} & \textbf{1} & \textbf{0} \\
\midrule
\multirow{2}{*}{\textbf{Actual}} & \textbf{1} & TP & FN \\
 & \textbf{0} & FP & TN \\
\bottomrule
\end{tabular}
\end{table}

\textbf{Accuracy} refers to the proportion of all judgments that are correct, i.e., judging the positive as positive and the negative as negative; the larger the value, the better:
\begin{equation}
    \mathrm{Accuracy} = \frac{TP+TN}{TP+TN+FP+FN}.
    \label{eq:acc}
\end{equation}
\textbf{Recall} is relative to the samples, i.e., how many of the positive samples are predicted correctly:
\begin{equation}
    \mathrm{Recall} = \frac{TP}{TP+FN}.
    \label{eq:recall}
\end{equation}
Similarly, \textbf{Precision} usually represents how many of the samples predicted as positive truly belong to the positive class:
\begin{equation}
    \mathrm{Precision} = \frac{TP}{TP+FP}.
    \label{eq:precision}
\end{equation}
To evaluate the comprehensive performance of a model under different confidence thresholds, Average Precision (AP) is introduced. Its essence is the numerical integration of the precision--recall (P--R) curve for a single class. In multi-class tasks, the mean Average Precision (mAP) is used as the evaluation standard, expressed as the arithmetic mean of the AP over all $N$ classes:
\begin{equation}
    \mathrm{mAP} = \frac{1}{N}\sum_{i=1}^{N} \mathrm{AP}_{i}.
    \label{eq:map}
\end{equation}
Commonly used dimensions include mAP@0.50 (computed only at IoU $=0.50$) and mAP@0.50:0.95 (the mean computed over the IoU interval from 0.50 to 0.95 with a step of 0.05, a stricter standard).

\section{Experiments}
\subsection{Experimental Setup}
The experimental environment is shown in Table~\ref{tab:env}. The key hyperparameters during model training are set as follows: the input image resolution is uniformly resized to $640\times640$, the batch size is set to 16, the maximum number of training epochs is 100, the initial learning rate is set to 0.01, the SGD optimizer is adopted with a momentum of 0.937 and a weight-decay coefficient of 0.0005, and mixed-precision training is enabled to accelerate convergence. For the training strategy, Early Stopping is used for regularization control: training is automatically terminated when the validation-set performance does not improve for 20 consecutive epochs, so as to ensure the generalization ability of the model.

\begin{table}[htbp]
\centering
\caption{Experimental environment.}
\label{tab:env}
\begin{tabular}{@{}p{6.2cm} p{6.2cm}@{}}
\toprule
\textbf{Hardware environment} & \textbf{Software environment} \\
\midrule
Processor: 13th-gen Intel Core i9-13900HX; Memory: 62.6\,GB & OS: Windows 11; Compute platform: CUDA 12.8; Deep-learning library: PyTorch \\
\bottomrule
\end{tabular}
\end{table}

\subsection{Datasets}
The datasets are described in Table~\ref{tab:datasets}. To comprehensively evaluate the effectiveness and generalization ability of the models generated by Cognitive-YOLO, we selected four scenarios with significantly different data distributions---rail-surface defects, rice-disease detection, fire detection, and student-behavior recognition---for cross-comparison testing.

\begin{table}[htbp]
\centering
\caption{Dataset description.}
\label{tab:datasets}
\begin{tabular}{@{}l p{8.5cm}@{}}
\toprule
\textbf{Dataset} & \textbf{Description} \\
\midrule
Rail Surface Defect & Industrial defect-detection dataset, used to detect defect types in the welding process. \\
Rice Disease & Agricultural pest/disease-detection dataset, used to identify common rice-growth disease types. \\
Fire Detection & Smoke- and fire-detection dataset, used for fire-warning tasks in indoor and outdoor scenes. \\
Student Behavior & Human-behavior detection dataset, used to detect students' behavioral states in the classroom. \\
\bottomrule
\end{tabular}
\end{table}

\subsection{Results and Analysis}
\textbf{Training process.} Table~\ref{tab:main} compares the Cognitive-YOLO-generated models with the baseline models. The comparison baselines cover the current mainstream lightweight object-detection models YOLOv8n, YOLOv11n, YOLOv12n, and YOLOv13n.

\begin{table*}[htbp]
\centering
\caption{Performance comparison of the models generated by Cognitive-YOLO against mainstream lightweight baselines across four cross-domain datasets. Within each dataset, the best result for each metric is highlighted in \textbf{bold}.}
\label{tab:main}
\small
\begin{tabular}{l l c c c c c}
\toprule
\textbf{Dataset} & \textbf{Algorithm} & \textbf{Params} & \textbf{Precision} & \textbf{Recall} & \textbf{mAP@0.5} & \textbf{mAP@.5:.95} \\
\midrule
\multirow{5}{*}{Rail Surface Defect}
& YOLOv8n  & 2.7M & 0.848 & 0.880 & \textbf{0.902} & 0.712 \\
& YOLOv11n & 2.6M & 0.908 & 0.835 & 0.878 & 0.648 \\
& YOLOv12n & 2.5M & 0.847 & 0.878 & 0.899 & \textbf{0.716} \\
& YOLOv13n & 2.5M & \textbf{0.911} & 0.842 & 0.882 & 0.660 \\
& \textbf{Cognitive-YOLO} & \textbf{1.9M} & 0.868 & \textbf{0.876} & \textbf{0.902} & 0.713 \\
\midrule
\multirow{5}{*}{Rice Disease}
& YOLOv8n  & 2.7M & 0.651 & \textbf{0.640} & 0.637 & 0.242 \\
& YOLOv11n & 2.6M & \textbf{0.719} & 0.532 & 0.608 & 0.220 \\
& YOLOv12n & 2.5M & 0.680 & 0.564 & 0.616 & 0.242 \\
& YOLOv13n & 2.7M & 0.649 & 0.711 & 0.587 & 0.226 \\
& \textbf{Cognitive-YOLO} & \textbf{2.1M} & 0.683 & 0.630 & \textbf{0.655} & \textbf{0.247} \\
\midrule
\multirow{5}{*}{Fire Detection}
& YOLOv8n  & 2.7M & 0.781 & 0.702 & 0.773 & 0.456 \\
& YOLOv11n & 2.6M & \textbf{0.796} & 0.711 & 0.779 & 0.462 \\
& YOLOv12n & 2.5M & 0.793 & 0.702 & 0.775 & 0.459 \\
& YOLOv13n & 2.7M & 0.754 & 0.689 & 0.753 & 0.437 \\
& \textbf{Cognitive-YOLO} & 2.6M & 0.782 & \textbf{0.718} & \textbf{0.783} & \textbf{0.466} \\
\midrule
\multirow{5}{*}{Student Behavior}
& YOLOv8n  & 2.7M & 0.887 & 0.876 & 0.929 & 0.769 \\
& YOLOv11n & 2.6M & 0.887 & 0.868 & 0.928 & 0.773 \\
& YOLOv12n & 2.5M & 0.882 & 0.877 & 0.926 & 0.777 \\
& YOLOv13n & 2.7M & \textbf{0.899} & 0.867 & 0.927 & 0.769 \\
& \textbf{Cognitive-YOLO} & 2.7M & 0.875 & \textbf{0.892} & \textbf{0.931} & \textbf{0.780} \\
\bottomrule
\end{tabular}
\end{table*}

The models generated by the system exhibit extreme lightweighting and efficient feature-extraction capability in complex-background and fine-grained feature scenarios. In the rice-disease detection task, because the background is complex and the objects are tiny, the baseline models generally show an imbalance between Precision (P) and Recall (R). In contrast, the model generated by our system, with only 2.1M parameters (the smallest in its group), achieves the highest comprehensive metric mAP@0.5:0.95 of 24.7\%, surpassing all fixed-architecture baseline models. In the rail-surface defect dataset, the system further compresses the parameter count to 1.9M (a significant reduction of about 24\% compared with YOLOv12n/v13n), while mAP@0.5 remains in the highest tier at 90.2\%. This fully demonstrates the superiority of the LLM-driven adaptive architecture in feature-expression efficiency, which is extremely suitable for resource-constrained edge-side deployment.

In scenarios with morphologically variable objects, the system achieves an accuracy breakthrough at the same compute scale. In the fire-detection and student-behavior recognition tasks, the parameter counts of the system-generated models (2.6M and 2.7M) are at the same magnitude as the baseline models. Under this premise, the system effectively utilizes the parameter capacity, and its mAP@0.5:0.95 metrics reach 46.61\% and 78.07\% respectively, both ranking first. This shows that when hardware constraints are relaxed, the adaptively generated architecture can approach a higher accuracy ceiling.

\begin{figure}[htbp]
   \centering
   \includegraphics[width=0.98\textwidth]{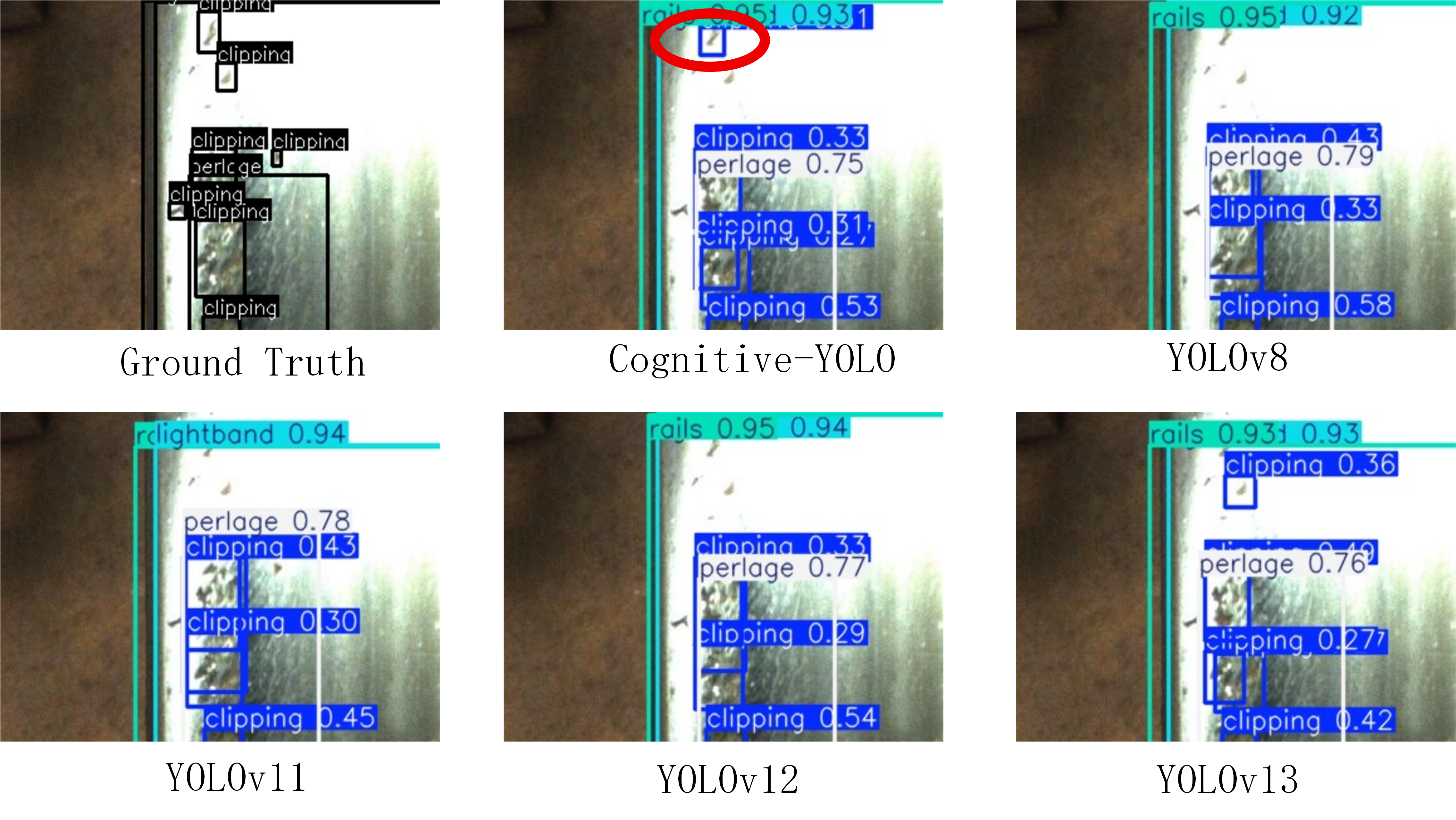}
   \caption{Detection-performance comparison on hard samples from the rail-surface defect dataset (top-left: ground-truth annotations). Cognitive-YOLO detects valid defects that the baseline models miss (one example highlighted with a red circle).}
   \label{fig:detect}
\end{figure}

\textbf{Hard-sample detection and visualization analysis.} As shown in Figure~\ref{fig:detect}, the model generated by the system can detect objects that the baseline models such as YOLOv8n, YOLOv11n, and YOLOv12n fail to detect, while the parameter count is reduced by about 20\% compared with the baseline models.

From the data distribution in Table~\ref{tab:detect}, it can be observed that, facing hard samples with severe interference, the fixed-architecture baseline models generally fall into the performance bottleneck of ``high missed detection.'' For example, although YOLOv8 and YOLOv12 maintain a high Precision, their missed-detection counts (FN) are as high as 20 and 22 respectively, resulting in a Recall below 0.48. This indicates that traditional baseline models tend to make conservative predictions in such scenarios, and their feature-mining ability is limited.

\begin{table}[htbp]
\centering
\caption{Detection-performance comparison on hard samples (38 ground-truth objects).}
\label{tab:detect}
\small
\begin{tabular}{l c c c c c c c}
\toprule
\textbf{Model} & \textbf{Total} & \textbf{Correct} & \textbf{Missed} & \textbf{False} & \textbf{Prec.} & \textbf{Rec.} & \textbf{F1} \\
\midrule
Ground truth & 38 & -- & -- & -- & -- & -- & -- \\
\midrule
\textbf{Cognitive-YOLO} & 24 & \textbf{19} & \textbf{19} & 5 & 0.792 & \textbf{0.500} & \textbf{0.613} \\
YOLOv8  & 21 & 18 & 20 & 3 & \textbf{0.857} & 0.474 & 0.610 \\
YOLOv11 & 22 & 15 & 23 & 7 & 0.682 & 0.395 & 0.500 \\
YOLOv12 & 19 & 16 & 22 & 3 & 0.842 & 0.421 & 0.561 \\
YOLOv13 & 21 & 17 & 21 & 4 & 0.810 & 0.447 & 0.576 \\
\bottomrule
\end{tabular}
\end{table}

The system-generated model successfully extracts the richest effective features, achieving the highest number of true-positive detections and the lowest missed-detection count among the strong baselines, and its Recall leads all baseline algorithms on the key targets. Among the 38 real defects, the system-generated model detects 24 objects, of which 19 are correctly detected; in contrast, YOLOv11 detects 22 objects but only 15 are correct, with the most missed detections. Compared with YOLOv11, the system-generated model correctly detects 4 more objects while missing 4 fewer.

On the F1-score metric, the system-generated model reaches 0.613, ranking first, with a clear advantage over YOLOv8's 0.610 and YOLOv11's 0.500. This benefits from its achieving the highest recall on the key targets while maintaining a relatively high precision (79.2\%). The system-generated model detects 24 objects on this sample, the most among all models. More importantly, its number of correct detections (19) also leads, which indicates that the additional objects it detects are mostly real and valid defects.

\begin{figure}[htbp]
   \centering
   \begin{subfigure}[b]{0.48\textwidth}
       \centering
       \includegraphics[width=\textwidth]{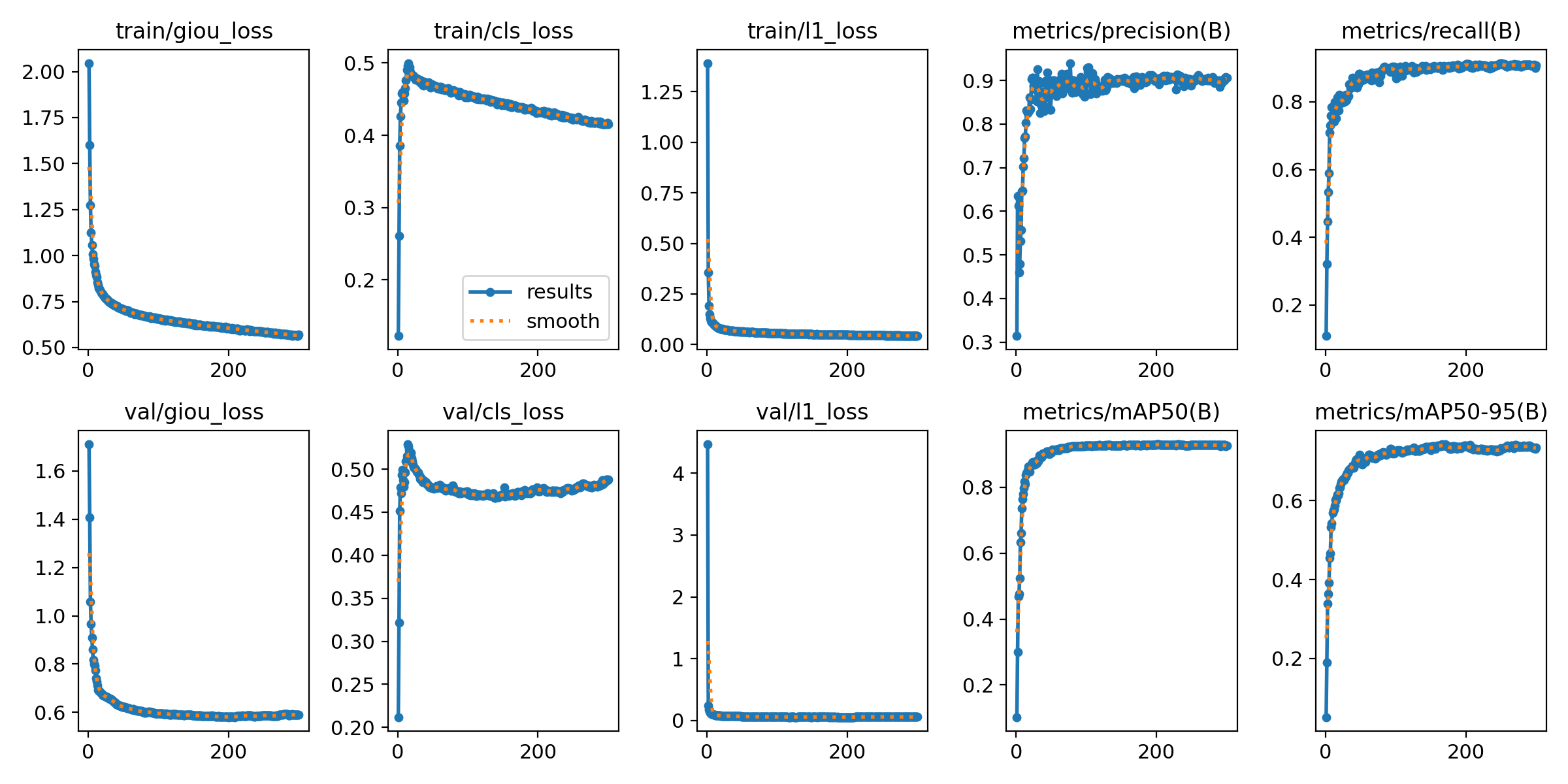}
       \caption{Cognitive-YOLO (ours)}
       \label{fig:train-ours}
   \end{subfigure}
   \hfill
   \begin{subfigure}[b]{0.48\textwidth}
       \centering
       \includegraphics[width=\textwidth]{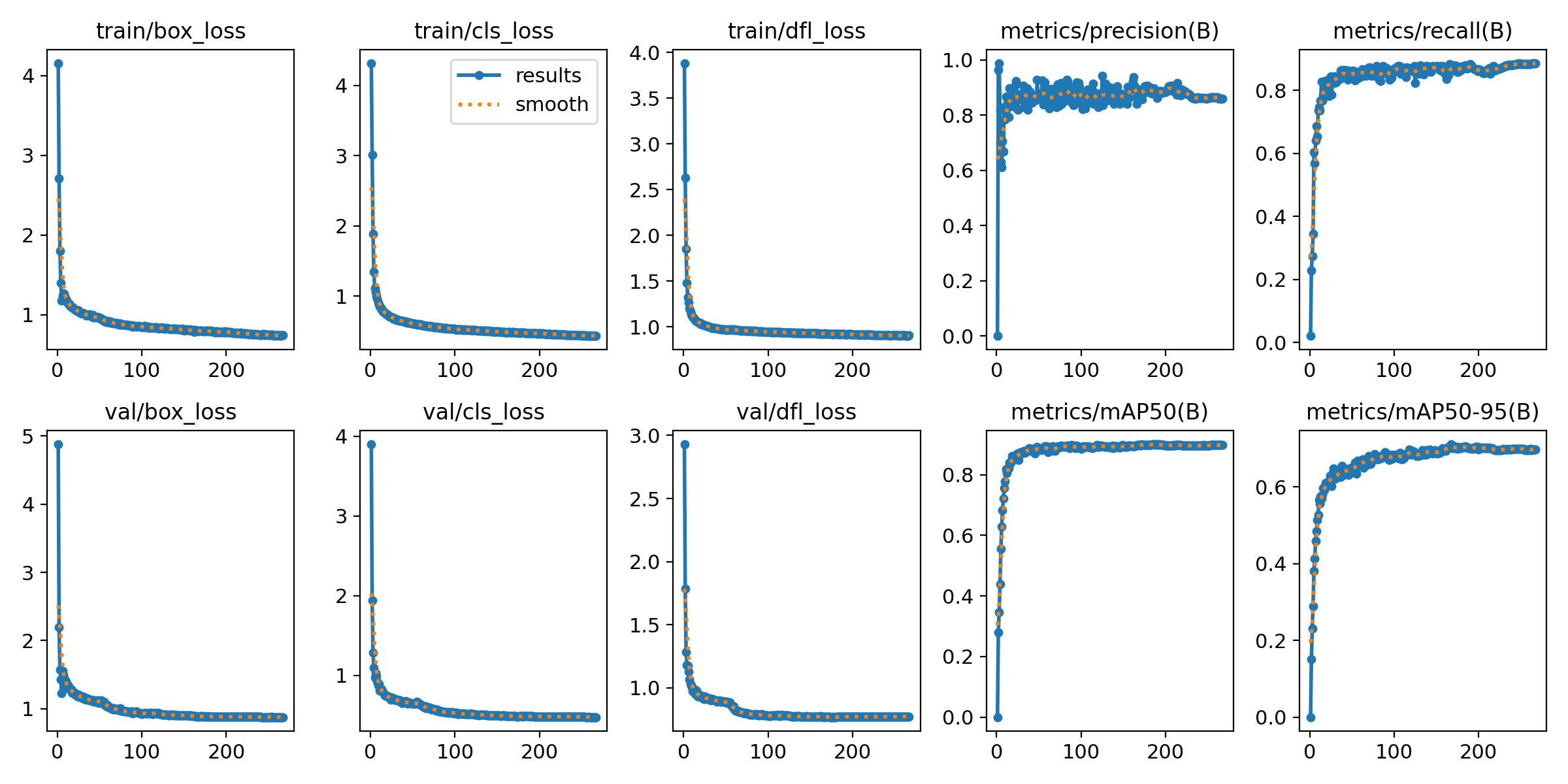}
       \caption{YOLOv11n (baseline)}
       \label{fig:train-base}
   \end{subfigure}
   \caption{Training-process metrics. The curves of Cognitive-YOLO are smooth and stable, whereas YOLOv11n exhibits an early gradient spike and persistent sawtooth oscillation in its precision curve.}
   \label{fig:train}
\end{figure}

\textbf{Training process and convergence-stability analysis.} Both the system-generated model and YOLOv11n adopt the early-stopping strategy during training. Figure~\ref{fig:train} shows the training-process metrics of the two models.

All the loss curves of the system-generated model (including the training-set and validation-set box\_loss, cls\_loss, and dfl\_loss) present a very standard, smoothly descending ``L''-shaped curve without any abnormal jitter, demonstrating a stable training process. In contrast, in the first few epochs of YOLOv11n's training, there is an obvious huge spike (the value rises to nearly 15), indicating severe gradient instability in the early training stage. Moreover, the precision curve of YOLOv11n is always accompanied by large sawtooth fluctuations in the later training process, indicating a certain uncertainty in its discrimination between positive and negative samples. In contrast, the Precision, mAP50, and mAP50-95 curves of the system-generated model show an almost perfect smooth upward trend. This shows that the architecture generated by the system is more consistent with the data distribution of the current dataset at the underlying design, and the model has learned more generalized and more solid deep semantic features.

\section{Conclusion}
With the rapid development of computer-vision technology, the application environment of object detection in vertical fields has become increasingly complex, and the demand for customization for specific scenarios (such as industrial quality inspection and security monitoring) continues to emerge. How to efficiently design high-precision, lightweight detection networks for different datasets has become an urgent problem to be solved. Although traditional neural architecture search (NAS) and manually designed networks have achieved remarkable results in performance, their heavy reliance on expert experience, high trial-and-error cost, and slow search process lead to an extremely long model-customization cycle. At the same time, real-world datasets often have extreme scale spans and class imbalance, causing fixed-architecture general detection models to have insufficient generalization ability in specific scenarios. The Cognitive-YOLO system designed and implemented in this paper, based on a large language model (LLM), effectively solves these problems. Nevertheless, the system still has some shortcomings and areas for improvement:

First, the SOTA module knowledge base in the current system is mainly built on currently known open-source network components, which has certain limitations. In the future, a large language model can be introduced to directly mutate and generate the underlying operator code, ensuring that the generated model architecture can break through the upper limit of existing modules.

Second, the experimental verification in this paper is mainly completed in a standard software-and-hardware experimental environment. In the future, it is necessary to further incorporate specific edge-side embedded hardware constraints (such as NPU and FPGA), and to bring hardware latency and energy consumption into the architecture-reasoning closed loop of the large model, so as to achieve more practical hardware-aware adaptive compilation.

\bibliographystyle{unsrt}
\bibliography{references}

\end{document}